\documentclass{article}
\usepackage{amsmath,graphicx,mlspconf}
\usepackage{xcolor}
\usepackage{amsmath}
\usepackage{amssymb}
\usepackage{amsfonts}
\usepackage{algorithm}
\usepackage{algorithmic}
\usepackage{booktabs}
\usepackage{url}
\usepackage{multirow}

\title{A NOVEL LATENT-CLASS ATTACK AND ITS DETECTION\\ BY CLASS SUBSPACE ORTHOGONALIZATION}

\name{%
    Guangmingmei Yang$^{\star}$
    \qquad David J. Miller$^{\star \dagger}$%
    \qquad George Kesidis$^{\star \dagger}$\thanks{This research supported
in part by NSF grant 2415752.}%
}
\address{
    $^{\star}$ CSE \& EE Depts, Penn State, University Park, PA, 16802\\
    $^{\dagger}$ Anomalee Inc., State College, PA, 16803
}

\begin{document}

\maketitle

\begin{abstract}
Deep learning, which in general relies on voluminous amounts of training data, is vulnerable to data poisoning attacks,
including error-generic attacks and backdoors (Trojans).  In this work, we propose a new data poisoning
attack we dub a {\it latent class} attack.  Here, all poisoned examples are from a class that is {\it novel} (unknown) for the given classification domain 
and are mislabeled to one of the known classes
(the target class) of the domain, so that the model learns to recognize the novel class as a sub-class of
the target class.  Such attacks could be used e.g. to defeat AI-based access control systems, or could cause a ``foe'' to be classified
as a ``friend''.  We also propose a post-training defense to detect this attack, {\it without any access to the training set}.  This
detection approach builds on ``class subspace orthogonalization'' (CSO), a plug-and-play paradigm demonstrated to improve existing backdoor 
detectors. Here, CSO is used to seek an input (a putative unknown class instance) whose internal representation is not aligned with any of the known classes, and yet which is 
classified with confidence to one of these classes.  Finally, specific to image classification domains, we propose a method for visualizing the estimated unknown
class instance, providing explainability to our latent class detections.
\end{abstract}

\newcommand{\cem}[1]{\textcolor{blue}{cem: #1}}
\section{Introduction}
\label{sec:intro}

Modern machine learning systems are increasingly trained on large-scale datasets whose quality and integrity are difficult to inspect exhaustively. This creates serious reliability and security risks.
Error-generic data poisoning attacks mislabel training samples with the goal of degrading the model's generalization accuracy.
Backdoor attacks alter training samples with subtle backdoor trigger patterns, and may mislabel these samples to a target class of the attack.  Then, operationally, if the trigger is present, a test sample will with high probability be misclassified to the target class.
In this paper, we propose a novel type of data poisoning attack we dub a {\it latent class attack}. Here, samples from an unknown/unrecognized category (a latent class) are introduced into the training set and mislabeled to a common {\it known} category -- the target class of the attack.  In this way, the model learns to classify latent class samples to the target category, i.e., to effectively treat the latent category as a sub-class of the target class.  Such attacks could be used e.g. to defeat AI-based access control systems, or could cause an unknown/zero-day ``foe'' class to be classified as a ``friend''. 
Unlike conventional backdoor attacks \cite{BadNets}, this setting does not involve an explicit pixel-level (or token-level) trigger or a trigger-insertion (prompt injection) mechanism. Instead, the model is trained to absorb an entire latent class into an existing known class.

The presence of undeclared classes in training or evaluation data has been studied in several related settings, including out-of-distribution detection, anomaly detection, novelty detection, and open-set recognition. However, these settings are primarily data-centric. Their goal is typically to reject unfamiliar test samples, identify abnormal inputs when they appear, or discover hidden groups from an available dataset. Similarly, some security works study label-based or semantic backdoor attacks in which only labels are modified during training. However, these attacks still rely on a trigger applied at test (operation) time.  In contrast, a latent-class attack does not require a trigger to activate the failure.

The problem of detecting such attacks 
is particularly challenging in the most realistic, {\it post-training} setting, wherein the defender does not have access to the original training data and can only inspect the trained model.  This setting applies e.g. when the defender lacks ``data rights'', i.e., when the training set is unavailable proprietary information, and it also applies when one is using or fine-tuning foundation models, for which the training set that was used is completely unknown.  
Even if a clean validation set is available for detection, the latent class (an {\it unknown} class) will be absent from it. As a result, the model can appear benign under standard accuracy-based evaluation while still mapping latent-class samples to the known class into which they were mislabeled. The defender must determine whether a trained model has absorbed an undeclared class into one of its known classes, without access to the poisoned training set and without prior knowledge of what the latent class ``looks like''. Beyond detection, it is also desirable to identify the target class and recover interpretable evidence about the latent class's learned features (visual ones, in the case of image classification).

To address this challenging detection problem, we propose \emph{Latent-class-CSO} (LC-CSO), a post-training defense method for latent-class attacks. Class Subspace Orthogonalization (CSO) \cite{cso} 
is a plug-and-play paradigm that has been demonstrated to improve the detection performance for a variety of backdoor detection methods.  
CSO is a 
general optimization approach wherein the detector's loss objective (e.g., aiming to find an input perturbation (a putative backdoor trigger) that induces samples to be classified to a putative target class of an attack) is altered to include a term penalizing internal feature directions that are not {\it orthogonal to} the intrinsic features of the putative target class.
CSO thus guides the detector to search in feature directions orthogonal to intrinsic features of the putative target class, where the backdoor trigger's features are most likely to ``reside''.
In this work, we adapt CSO to 
suppress features aligned with {\it all known-class subspaces} while searching for an input that is, {\it with high confidence}, classified to one of the known categories. Intuitively, a clean model should not contain a high-confidence class direction outside the subspaces explained by its known classes. In contrast, a model trained with latent-class poisoning may contain such a direction: the latent class has been absorbed into the target decision region but is not fully explained by the intrinsic target-class features.

To avoid poor local minima, we minimize our LC-CSO objective using multiple random restarts, with the detection decision based on the best solution amongst this set.  The high-confidence, decided-to class is an estimate of the target class of the attack.  Moreover, the optimized input, if a detection is made, is an estimate of an instance of the latent class.  Thus, we also propose a companion method for coherently visualizing (for image classification domains) these estimated latent class instances.
These visualizations are not recovered training images; rather, they can provide interpretable evidence about the latent class or the feature directions introduced by the latent-class poisoning.

Our contributions are two-fold. First, we formalize the \emph{latent-class attack}, a data-poisoning setting in which samples from an undeclared class are inserted into the training set and labeled as a known target class. This setting differs from conventional backdoor attacks because no explicit trigger is inserted and no source-class trigger mechanism is required. Second, we propose \emph{LC-CSO}, a post-training defense that detects latent-class poisoning, identifies the target class, and produces interpretable inversion evidence of the latent-class features. By optimizing for high-confidence predictions outside the intrinsic subspaces of known classes, LC-CSO exposes representational structure that is absent in clean models but induced by latent-class absorption.

\section{Related Work}
\label{sec:related_works}

Our setting is related to several lines of work, but differs in both the threat model and defender's data access assumptions.

\textbf{Label-based poisoning and semantic backdoors.}
Several prior works study attacks that modify training labels or exploit semantic attributes. For example, \cite{294500} proposes a semantic backdoor attack in which source-class samples sharing certain semantic attributes with the target class are relabeled, causing the model to associate them with the target class. Similarly, \cite{jha2023label} studies a label-only backdoor attack that flips the labels of carefully selected training samples. Although no explicit pixel pattern is inserted into the poisoned images during training, the selected samples are chosen so that their training dynamics resemble those of conventional backdoor poisoning; these attacks still depends on a trigger-like condition applied at test time.
Note that our attack fundamentally differs from these in that it does not require a trigger.  Moreover, our attack introduces a wholly
unknown class into a given classification domain.

\textbf{Out-of-distribution detection and open-set recognition.}
A second related line of work is out-of-distribution (OOD) detection, which studies whether a test input comes from a distribution different from the one observed during training. OOD detection broadly includes anomaly detection, novelty detection, open-set recognition, and related ``unknown unknown'' settings \cite{yang2024generalized}. These methods usually address a sample-level decision problem: given a test input, determine whether it should be accepted as in-distribution or rejected as novel, anomalous, or open-set. This perspective is relevant because the latent class in our setting lies outside the declared label space.
More specifically, anomaly detection and novelty detection aim to identify samples that deviate from a predefined notion of normality or that do not belong to any known training category. Open-set recognition considers test samples from classes that are absent during training. The ``unknown unknowns'' setting of \cite{zhao2024exploratory} is particularly related, since it considers unknown classes that are present in the training data but are misperceived as known labels under the available supervision. However, its goal is to inspect the training data and discover hidden classes.

Our detection setting differs from this prior work in two key ways. First, we do not assume that the defender has access to the possibly poisoned training set. Second, we do not have access to instances of the latent class at test time because we consider a post-training (and before deployment) 
model inspection problem: given only a trained classifier and a small clean dataset from the known classes, determine whether the model contains evidence that an undeclared latent class was absorbed into one of the known classes during training. Thus, while OOD detection asks whether a particular input is unfamiliar to the model, our problem asks whether the model itself encodes evidence of a hidden mislabeled class.

Clustering is also frequently used in OOD, open-set, and unknown-class discovery settings \cite{shu2020p,liu2020few,chen2020learning,yang2021semantically}. Given access to a dataset, one can cluster samples in input, representation, or prediction space to identify natural groups, potentially revealing subpopulations within a labeled class. However, the existence of a cluster does not by itself establish that the subgroup is a valid semantic class, nor does it determine whether the model was trained on a mislabeled latent class. Moreover, clustering-based approaches are inherently data-centric: they require access to samples containing the hidden group. In contrast, our goal is model-only post-training inspection without access to the training set or to latent-class samples. When such data are available, our method could be complementary to clustering; for example, a CSO-based statistic could be computed for each sample in a discovered cluster and aggregated as a detection score. However, such a hybrid approach still requires access to the relevant data, whereas our primary setting does not.

\textbf{Class Subspace Orthogonalization.}
We adapt CSO to the latent-class setting with a different goal from \cite{cso}. Rather than suppressing only the intrinsic features of a hypothesized target class, we suppress the intrinsic feature subspaces of \textit{all known classes}. This forces the optimization to search for high-confidence decision directions that are not well explained by the clean known-class feature subspaces. For a clean model, it should be difficult to find such directions. In contrast, if a latent class has been absorbed into a target class, the model may contain ``residual'' feature directions associated with the latent class that still lead confidently to a target class prediction.

In summary, our problem is distinct from existing work because the latent class is not a trigger, no source-class trigger insertion is assumed, and the defender may not have access to either the poisoned training data or to latent-class test samples. We therefore study a model-only post-training inspection problem: given only a trained classifier and a small clean dataset from the known classes, determine whether an undeclared latent class has been absorbed into a known class, identify the affected target class, and extract interpretable evidence about the latent class's learned visual features.

 \section{Method}

\subsection{Latent-Class Attack}

A latent-class attack is a data-poisoning attack in which poisoning occurs through open-set mislabeling: samples from an undeclared class are assigned to a known target class, causing the trained model to absorb the latent class into the target decision region. We consider a setting in which the declared task contains \(K\) known classes,
$\mathcal{Y}_{\mathrm{known}}=\{1,\dots,K\}$.
In addition to these known classes, there exists an undeclared latent class, denoted \(\ell\),
$\ell \notin \mathcal{Y}_{\mathrm{known}}$.
The classifier is trained to predict only the \(K\) known classes.

Let \(t\in\mathcal{Y}_{\mathrm{known}}\) denote the target class into which latent-class samples are mislabeled. Let \(\mathcal{D}_{\mathrm{known}}\) denote the clean training data from the known classes, and let 
\(\mathcal{D}_{\ell}\subset\mathcal{D}_{\mathrm{known}}\) 
denote the available samples from the latent class $\ell\not\in\mathcal{Y}_{\mathrm{known}}$. Given a poisoning subset \(\mathcal{P}_{\ell}\subseteq \mathcal{D}_{\ell}\), the poisoned training set 
\[
\mathcal{D}_{\mathrm{train}}
=
\underbrace{
\{(x,y)
\in \mathcal{D}_{\mathrm{known}} \times \mathcal{Y}_{\mathrm{known}}
\}
}_{\text{clean known-class samples}}
\cup
\underbrace{
\{(x,t):x\in \mathcal{P}_{\ell}\}
}_{\text{latent-class samples}} .
\]
The poisoning rate (PR) is
$|\mathcal{P}_{\ell}|/|\mathcal{D}_{known}|$.

Let the trained classifier be
\[
f_{\theta}(x)=S_b(S_a(x))\in\mathbb{R}^{K},
\]
where \(\theta\) denotes all learned model parameters, \(S_a(\cdot)\) is the feature extractor, and \(S_b(\cdot)\) is the classifier head. The output \(f_{\theta,k}(x)\) denotes the logit for known class \(k\in\mathcal{Y}_{\mathrm{known}}\). The model never observes \(\ell\) as a valid label and can only predict one of the known classes in \(\mathcal{Y}_{\mathrm{known}}\).

The attack succeeds if held-out latent-class samples are mapped to the target
class, with the Attack Success Rate 
\[
\operatorname{ASR}
=
\frac{1}{|\mathcal{D}_{\ell}^{\mathrm{test}}|}
\sum_{x\in \mathcal{D}_{\ell}^{\mathrm{test}}}
\mathbf{1}
\left[
\arg\max_{k\in\mathcal{Y}_{\mathrm{known}}} f_{\theta,k}(x)=t
\right].
\]

\subsection{All-Class CSO Penalty}
For each known class \(k\in\mathcal{Y}_{\mathrm{known}}\), let
$\mathcal{D}^{test}_k=\{x^{(k)}\}$
be a small clean reference set that is available to the defender. Following \cite{cso}, we learn a class-specific soft mask
$v_k\in[0,1]^d$
that identifies the intrinsic feature components of class \(k\). The mask is learned by preserving the class prediction under the masked features while suppressing the prediction under the complementary features:
\[
\begin{aligned}
\min_{v_k}\quad
\sum_{x^{(k)}\in\mathcal D_k^{\text{test}}}
\Big[
&
\mathcal L\!\left(
S_b(S_a(x^{(k)})\odot v_k),\,k
\right)
\\
&-
\mathcal L\!\left(
S_b(S_a(x^{(k)})\odot(1-v_k)),\,k
\right)
\Big].
\end{aligned}
\]
where \(S_a(\cdot)\) is the feature extractor, \(S_b(\cdot)\) is the classifier head, \(\odot\) denotes element-wise multiplication, and \(\mathcal{L}\) is the cross-entropy loss.

For each known class \(k\), we define its masked intrinsic feature centroid as
\[
\mu_k
=
\frac{1}{|\mathcal{D}^{test}_k|}
\sum_{x^{(k)}\in\mathcal{D}^{test}_k}
v_k\odot S_a(x^{(k)}).
\]
This centroid summarizes the intrinsic benign feature direction of class \(k\).

Let \(z\in\mathcal{X}\) denote an input optimized by the inversion procedure. In the original CSO formulation, for a putative target class \(t\), the CSO penalty measures the positive cosine alignment between the optimized feature \(S_a(z)\) and the intrinsic features of class \(t\):
{\small
\[
C_t(z)
=
\frac{1}{|\mathcal{D}^{test}_t|}
\sum_{x^{(t)}\in\mathcal{D}^{test}_t}
\operatorname{ReLU}
\left(
\frac{
\left\langle
S_a(z),
v_t\odot S_a(x^{(t)})
\right\rangle
}{
\left\|S_a(z)\right\|_2
\left\|v_t\odot S_a(x^{(t)})\right\|_2
}
\right).
\]
}
This penalty is then added to the baseline detector objective for each putative target class.

In our setting, we instead define an all-class CSO penalty that measures the positive alignment between \(S_a(z)\) and the intrinsic feature centroids of all known classes:
\[
C(z)
=
\frac{1}{K}
\sum_{k\in\mathcal{Y}_{\mathrm{known}}}
\operatorname{ReLU}
\left(
\frac{
\left\langle
S_a(z), \mu_k
\right\rangle
}{
\left\|S_a(z)\right\|
\left\|\mu_k\right\|
}
\right).
\]
Penalizing \(C(z)\) while seeking high-confidence prediction to one of the known classes encourages the optimized input to move away from the intrinsic feature directions of all known classes. For a clean model, after these known-class directions are suppressed, there should be no residual feature direction that gives confident classification to one of the known classes. In contrast, if a latent class has been absorbed into a target class, the model may contain residual latent-class feature directions that are not explained by the clean known-class subspaces but still map confidently to a known class (the target class).

\subsection{LC-CSO}

In a latent-class attacked model, we expect that the optimized inputs can still be confidently
  classified into a common known class while known-class intrinsic features are being suppressed. This common
  class should correspond to the target class into which the latent class was absorbed. In contrast,
  for a clean model, the optimized inputs should yield lower-confidence predictions (or higher CSO penalties for a given level of confidence).

The softmax posterior over class $k\in\mathcal{Y}_{known}$ at input $z$,
  $$
  p_k(z) = \operatorname{softmax}_k\bigl(f_\theta(z)\bigr) = \frac{\exp\bigl(f_{\theta,k}(z)\bigr)}{\sum_{j\in\mathcal{Y}_{known}}\exp\bigl(f_{\theta,j}(z)\bigr)}.
  $$

  We can then assess the confidence with which $z$ is classified to one of the known classes via the Shannon entropy:
  $$
  H\bigl(z\bigr) = -\sum_{k\in\mathcal{Y}_{known}} p_k(z) \log p_k(z).
  $$

  The LC-CSO objective is, accordingly, then given as:
  $$
  \mathcal{L}(z) = \Bigl[\underbrace{C(z)}_{\text{CSO penalty}}\ +\ \beta \underbrace{H\bigl(z\bigr)}_{\text{Decision uncertainty}}\Bigr],
  $$
  where $\beta \geq 0$ controls the strength of the confidence term. 

Minimizing the CSO objective from a single random initialization can be unstable due to local optima and initialization effects. We therefore apply multiple random restarts, optimizing the LC-CSO objective starting from \(M\) restart initializations,
$z_1,\dots,z_M$.

  After optimization, for each restart $m$ define the per-restart CSO score $c_m = C(z_m)$, confidence
  $\pi_m = \max_{k}p_k^{(m)}$, and predicted class $\hat{y}_m = \arg\max_k p_k^{(m)}$. Detection and 
  target-class identification are derived from the single most off-manifold restart that achieves high 
  confidence: with $\mathcal{Q} = \{m:\pi_m \geq \tau_{\mathrm{conf}}\}$, 
  $
  m^\star = \arg\min_{m\in\mathcal{Q}} c_m, $ $q(\theta)=\min_{m\in\mathcal{Q}} c_m ~(=+\infty\text{ if }\mathcal{Q}=\varnothing),
  $
  a latent-class attack is indicated when this minimum CSO at high confidence is small relative to a
  reference threshold:
  $$
  \mathrm{flag}(\theta) = \mathbf{1}\bigl[q(\theta) < \tau_{\mathrm{cso}}\bigr]
  $$
  $$\hat{t} =
  \hat{y}_{m^\star}
  $$

  Intuitively, the attack creates a feasible direction in input space that is weakly aligned with
  known-class intrinsic features, yet still maps consistently to the absorbed target class. Such a
  direction should not exist in a clean model.
  So, for a clean model it is hypothesized no
  restart will achieve both low CSO and high confidence simultaneously, and $q(\theta)$ remains large.

\subsection{Fourier-Parameterized Inversion}

After detection with LC-CSO, we seek interpretable evidence about what the latent class may look like. Since the optimized inputs associated with a flagged target class are intended to expose residual feature directions not explained by the known-class intrinsic subspaces, they can be used as inversion-style visualizations of the latent-class-related features. However, directly optimizing an input $z$ in pixel space often produces high-frequency adversarial artifacts that are difficult to interpret.

To improve visual interpretability, we parameterize each optimized input in the Fourier domain rather than directly in pixel space. Let \(s\) denote the learnable complex spectrum associated with the best re-start. The corresponding image is generated by
\[
z
=
\sigma\!\left(
\text{FFT}^{-1}(s\odot \kappa)
\right),
\]
where $\text{FFT}^{-1}(\cdot)$ is the inverse fast Fourier transformation, \(\sigma(\cdot)\) maps pixel values to \([0,1]\), and \(\kappa\) is a frequency-dependent decay factor defined by
\[
\kappa_{u,v}
=
\frac{1}{
\max\!\left(
\rho_{u,v},
\frac{1}{\max(H,W)}
\right)^{\alpha}
},
\qquad
\rho_{u,v}
=
\sqrt{f_u^2+f_v^2}.
\]
Here, \(\alpha>0\) is a hyperparameter controlling the degree of spectral decay. Larger values of \(\alpha\) suppress high-frequency components more strongly, thereby producing smoother and more low-frequency-dominated images.

Importantly, the LC-CSO objective is unchanged. We still optimize the same objective,
but now $z$ is generated from its Fourier-domain variable $s$. Equivalently, the optimization is performed over $s$ through the composite mapping
\[
z = z(s)
=
\sigma\!\left(
\text{FFT}^{-1}(s\odot \kappa)
\right).
\]
Thus, the inversion problem becomes
\[
\min_{s}
\mathcal{L}\bigl(z(s)\bigr).
\]

The resulting optimized image \(z\) is used as a post-hoc inversion output. It provides interpretable evidence of the visual features or feature directions associated with the absorbed latent class after known-class intrinsic features have been suppressed.

\section{Experimental Results}
\label{sec:ExperimentalResults}
\subsection{Experiment Setup}

\textbf{Dataset and models.} We experiment on two benchmark datasets: CIFAR-10 
\cite{cifar10} and (a subset of) TinyImageNet \cite{TinyImageNet}, containing 10 classes. We evaluate our methods for ResNet-18 \cite{ResNet} and ResNet-34 on TinyImageNet. We randomly select 30 clean test samples per class for  LC-CSO.

\textbf{Attack settings.}
For each attacked model, we select one class as the latent class and one known class as the target class. All training samples from the latent class are relabeled as the target class. On CIFAR-10, we consider all ordered latent-class/target-class pairs, yielding 90 attacked models. On Tiny-ImageNet, we use a 10-class subset and similarly evaluate all ordered latent-class/target-class pairs, yielding another 90 attacked models. 
For each dataset, we also train 10 clean reference models. In each clean model, one class is excluded from both training and testing, so that the classifier is trained only on the remaining nine known classes without any latent-class relabeling.

\textbf{Defense settings.}
For LC-CSO, we use 30 randomly selected test samples per known class as the available clean set, i.e.,
\(
|\mathcal{D}^{test}_k| = 30, \forall k\in\mathcal{Y}_{\mathrm{known}}.
\)
We use the output of the last convolutional layer as \(S_a(\cdot)\), and set the consistency weight to \(\beta=0.0005\). We set the number of restarts $M=64$.
Since there are no existing defenses specifically designed for latent-class attacks, we compare against representative backdoor defenses, including Neural Cleanse (NC) \cite{NC} and MMBD \cite{MMBD}.

\textbf{Evaluation metrics.}
We evaluate each attack using the attack success rate (ASR) and clean test accuracy (ACC).

For defense, we evaluate each detector using ROC AUC. For attacked models, a detection is counted as correct only if the method both flags the model as compromised and correctly infers the target class. The clean models are used to evaluate the false positive rate.

\subsection{Latent Class Attack}
The performance of latent-class attacks is reported in Table \ref{tab:attack_performance}. All metrics are averaged over the attacked models for each dataset. For reference, clean ResNet-18 models achieve an average accuracy of 93.5\% on CIFAR-10, while clean ResNet-34 models achieve an average accuracy of 74.0\% on the Tiny-ImageNet subset. The latent-class attacks achieve high ASR while maintaining clean accuracy close to that of the corresponding clean models. So, the attack successfully maps latent-class samples to the target class without causing a substantial degradation in standard test accuracy.
\begin{table}[t]
\centering
\caption{Attack performance on CIFAR-10 \& TinyImageNet.}
\label{tab:attack_performance}
\begin{tabular}{ccc}
\toprule
Dataset & ASR & ACC\\
\midrule
CIFAR-10 &  93.4\% & 93.0\% \\
TinyImageNet &  77.2\% & 72.6\% \\
\bottomrule
\end{tabular}
\end{table}

\subsection{LC-CSO Detection Performance}
In Table \ref{tab:detection_performance}, we report the performance of LC-CSO on the latent class attack along with baselines NC and MMBD on CIFAR-10 and TinyImageNet subset. We compute AUC using 90 attacked models and 10 clean models for each dataset. LC-CSO outperforms both baselines on both datasets, demonstrating that latent-class attacks require a different detection signal from conventional trigger-based backdoor defenses.
\begin{table}[t]
\centering
\caption{Detection ROC AUC on CIFAR-10 \& TinyImageNet.}
\label{tab:detection_performance}
\begin{tabular}{lcc}
\toprule
{Method} 
& {CIFAR-10} 
& {TinyImageNet} \\

\midrule
NC & 0.60 & 0.36 \\
MMBD & 0.58 & 0.60 \\
LC-CSO & 0.89 & 0.77 \\
\bottomrule
\end{tabular}
\end{table}

\subsection{LC-CSO Inversion Performance}
\begin{figure}
    \centering
    \includegraphics[width=1\linewidth]{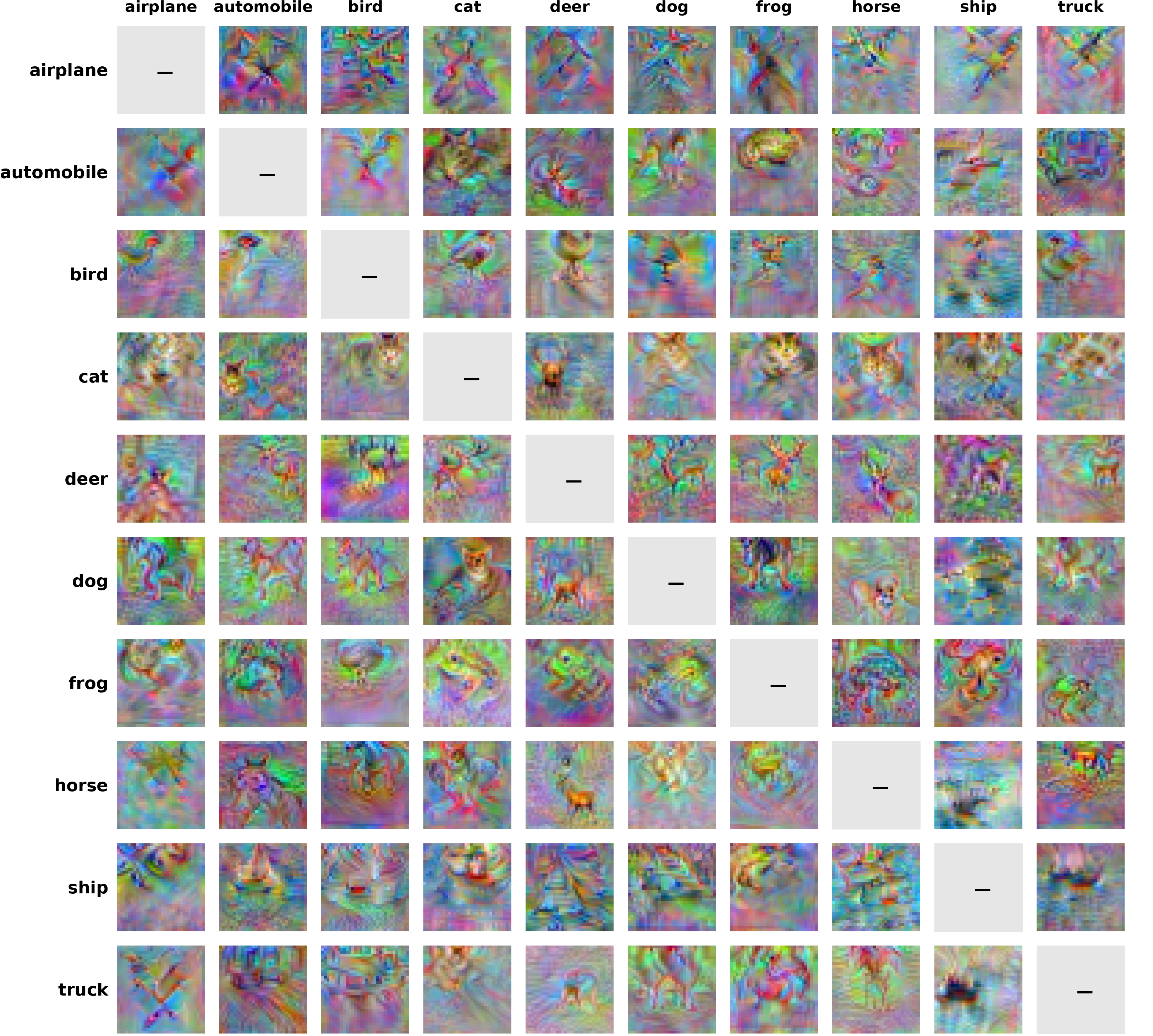}
    \caption{LC-CSO inversions on CIFAR-10. Rows and columns 
    respectively are the latent class and the target class.}
    \label{fig:cifar_inverse}
\end{figure}

In this section, we demonstrate the inversion images from LC-CSO. Figure \ref{fig:cifar_inverse}
shows the inverted images for all latent-class/target-class pairs on CIFAR-10. Across target classes, the recovered images consistently reveal recognizable visual characteristics of the corresponding latent class. For example, models trained with airplane as the latent class often produce cross-shaped structures resembling airplane bodies and wings. Similarly, the inversions capture cat-like facial patterns, deer-like horns and silhouettes, frog-like green textures, and triangular sail-like structures for ship. These results suggest that LC-CSO does not merely identify the affected target class, but can also provide interpretable visual evidence of the latent class features absorbed by the model.

\subsection{Ablation Studies}
\textbf{Centroid-based CSO.}
We compare our centroid-based CSO with the original per-sample CSO formulation \cite{cso}. On CIFAR-10, per-sample CSO achieves an AUC of 0.84, while centroid-based CSO improves the AUC to 0.89. This suggests that, for latent class attacks, per-sample CSO may over-penalize known class subspaces.

\textbf{Choice of $\beta$.}
We evaluate values of $\beta$: 0, 0.0001, 0.0003, 0.0005, 0.001 and 0.01. The corresponding AUC scores are 0.85, 0.86, 0.87, 0.89, 0.85, and 0.81.  We chose the $\beta$ value that achieves the lowest CSO penalty after the optimization is applied.

\section{Summary}
This work introduces a new data poisoning threat called a \emph{latent class attack}, where samples from an unknown class outside the declared classification domain are mislabeled as a known target class during training. As a result, the model learns to treat the unknown class as a subclass of the target class, which can create serious security risks. The paper also proposes a post-training defense that requires no access to the original training data. The defense adapts class subspace orthogonalization to search for inputs whose internal representations are not aligned with any known class, yet which are still confidently classified as one of them. For image classifiers, the method additionally provides visualizations of the inferred unknown class, giving interpretability to detections.

\bibliographystyle{IEEEbib}

\end{document}